%% file: iclr2027_conference.tex
\documentclass{article} % For LaTeX2e
\usepackage{iclr2027_conference,times}

\input{math_commands.tex}

\usepackage{url}
\usepackage{graphicx}
\usepackage{booktabs}
\usepackage{tabularx}
\usepackage{amsmath}
\usepackage{amssymb}
\usepackage{amsthm}
\usepackage{multirow}
\usepackage{enumitem}
\usepackage{algorithm}
\usepackage{algpseudocode}
\usepackage{xspace}
\usepackage{placeins}
\usepackage{float}
\usepackage{hyperref}

\graphicspath{{figures/}{../figures/}}

\newcommand{\SROPD}{\textsc{SR-OPD}\xspace}

\newcommand{\apptablestyle}{%
  \setlength{\tabcolsep}{2pt}%
  \renewcommand{\arraystretch}{1.05}%
}
\title{Spend Teacher Tokens Where They Matter:\\Success-Referenced On-Policy Distillation}

\author{%
Xiang Chen$^{1,4,*}$, Futao Su$^{3,*}$, Kong Wang$^{2}$\\
\bfseries Jiayi Chen$^{4}$, Tanlin Li$^{4,\dagger}$, Ziyu Su$^{4}$\\[0.5em]
\normalfont $^{1}$Tongji University\quad $^{2}$Hunan University\\
\normalfont $^{3}$Shanghai Jiao Tong University\quad $^{4}$Xiaohongshu\\[0.4em]
\normalfont\small $^{*}$Equal contribution (co-first authors).\quad $^{\dagger}$Corresponding author.
}

\begin{document}

\maketitle

\input{sections/00_abstract}
\input{sections/01_intro}
\input{sections/02_related}
\input{sections/03_method}
\input{sections/07_experiments}
\input{sections/08_conclusion}
\input{sections/09_statements}

\bibliography{iclr2027_conference}
\bibliographystyle{iclr2027_conference}

\appendix
\setcounter{table}{0}
\renewcommand{\thetable}{\Alph{section}\arabic{table}}
\renewcommand{\theHtable}{appendix.\Alph{section}.\arabic{table}}
\setcounter{figure}{0}
\renewcommand{\thefigure}{\Alph{section}\arabic{figure}}
\renewcommand{\theHfigure}{appendix.\Alph{section}.\arabic{figure}}
\input{sections/10_appendix}

\end{document}

%% file: math_commands.tex
\usepackage{amsmath,amsfonts,bm}

\def\eqref#1{(\ref{#1})}
\def\1{\bm{1}}

\DeclareMathAlphabet{\mathsfit}{\encodingdefault}{\sfdefault}{m}{sl}
\SetMathAlphabet{\mathsfit}{bold}{\encodingdefault}{\sfdefault}{bx}{n}

%% file: sections/00_abstract.tex
\begin{abstract}
On-policy distillation (OPD) combines student-generated rollouts with dense token-level supervision from a teacher, but providing such supervision for every rollout requires substantial teacher computation. We introduce Success-Referenced On-Policy Distillation (SR-OPD), which reduces this cost by selecting which prompts and rollouts receive teacher supervision. When the student produces both successful and failed rollouts for the same prompt, a successful rollout can serve as a natural reference for selecting failed rollouts. SR-OPD therefore focuses on such prompts and prioritizes failed rollouts whose hidden-state trajectories show sustained divergence from a successful reference, while accounting for estimated teacher-input cost. Across three teacher--student pairs and six mathematical reasoning benchmarks, SR-OPD uses only 3.46--5.02\% of the teacher-input tokens required by Vanilla OPD in the one-pass setting while maintaining comparable reasoning performance. Under a controlled setting matched to 5\% of Vanilla OPD's teacher-input budget, further experiments support both key design choices: focusing supervision on prompts with both successful and failed rollouts, and using successful rollouts to guide failure selection. These results indicate that a student's own successful behavior can serve as a useful reference for allocating teacher supervision under a fixed teacher-input budget.
Code is available at \url{https://github.com/ChenXiangYpily1234/SR-OPD}.
\end{abstract}

%% file: sections/01_intro.tex
\section{Introduction}
\label{sec:intro}

On-policy distillation (OPD) trains a student on rollouts generated by its current policy using supervision from a teacher model. By aligning training with the student's inference-time rollout distribution, OPD mitigates the train--test mismatch of offline distillation~\citep{agarwal2024gkd,gu2024minillm}. However, querying the teacher for every rollout incurs substantial cost. Such uniform supervision can be inefficient because the value of teacher supervision varies across prompts, rollouts, and token positions, making uniform allocation of teacher computation undesirable under a fixed teacher-input budget. This raises a natural question: can the student decide, before teacher inference, which prompts and rollouts should receive teacher supervision?

Recent work improves OPD efficiency through selective supervision. Methods such as TIP~\citep{xu2026tip}, TA-OPD~\citep{wang2026taopd}, and SelecTKD~\citep{huang2025selectkd} identify or reweight informative token-level supervision, improving the use of teacher signals within student rollouts. However, token-level selection does not by itself avoid the teacher-input cost of processing an entire queried rollout. A closely related direction is TLR~\citep{anonymous2026tlr}, which performs pre-query rollout routing in OPSD, where supervision is provided by a privileged-conditioned version of the same model rather than a separate teacher model. Thus, pre-query routing for standard teacher--student OPD remains underexplored.

\begin{figure}[h]
\centering
% Editable source: ../SR_OPD_Routing_Reference_Fonts_Compact_Layout_v3.pptx
\includegraphics[width=0.99\textwidth]{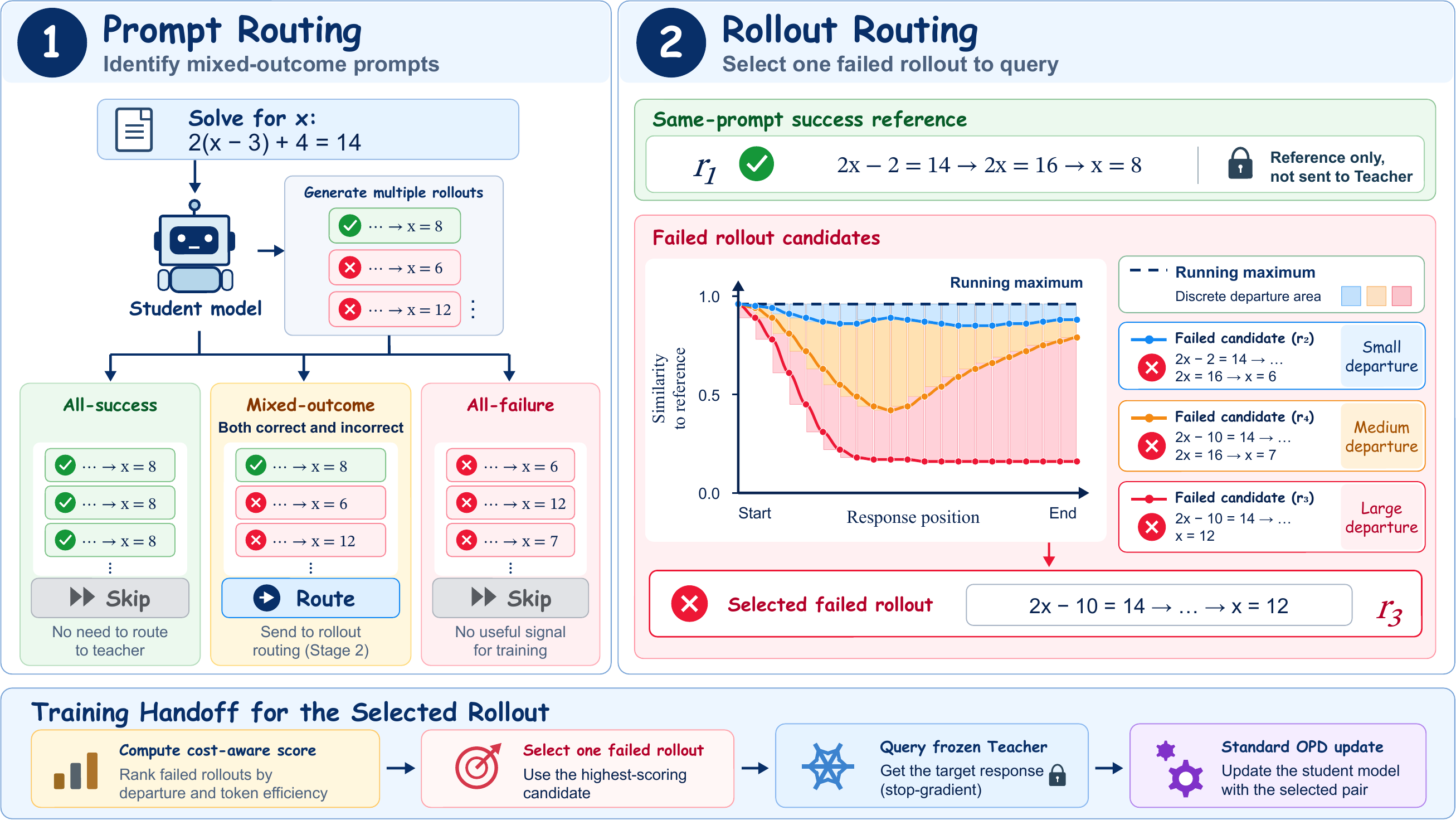}
\caption{Two-stage pre-query routing in SR-OPD. Left: prompt routing retains prompts containing both successful and failed rollouts while skipping all-success and all-failure prompts. Right: a same-prompt success serves only as a student-side reference; persistent-departure area (PDA) and teacher-input cost rank failed rollouts. In this example, only $r_3$ receives frozen-teacher supervision and the standard OPD update.}
\label{fig:method_overview}
\end{figure}

To address this gap, we introduce \textbf{Success-Referenced On-Policy Distillation (SR-OPD)}, a two-stage pre-query routing method illustrated in Figure~\ref{fig:method_overview}. A similar filtering principle has been used in on-policy reinforcement learning, for example in DAPO's dynamic sampling~\citep{yu2025dapo}. In SR-OPD, however, this filtering serves a different purpose: prompts for which the student produces both successful and failed rollouts provide failed rollouts to supervise and a same-prompt successful rollout that can serve as a teacher-free reference. The first stage, \textbf{prompt routing}, therefore retains such prompts while skipping all-success prompts, which contain no failed rollout to target, and all-failure prompts, which lack a successful reference for the second stage. This reduces teacher-query cost while providing the successful reference required for subsequent rollout routing.

The second stage, \textbf{rollout routing}, uses the same-prompt successful rollout as a reference for selecting among failed rollouts. We introduce \textbf{persistent-departure area (PDA)}, a teacher-free measure of sustained representational divergence between successful and failed reasoning rollouts. Specifically, PDA captures persistent departure of a failed rollout's hidden-state trajectory from the successful reference: temporary deviations that later recover yield small values, whereas sustained departures yield large ones. SR-OPD combines PDA with estimated teacher-input cost to prioritize failed rollouts and queries the teacher only for the highest-scoring failed rollout. The successful rollout is used solely as a student-side reference and is never included in the teacher context. After routing, the selected rollout receives the standard OPD target and update, leaving the underlying OPD procedure unchanged.

Our contributions are threefold:
\begin{enumerate}[leftmargin=*,itemsep=1pt,topsep=1pt]

\item \textbf{Pre-query routing for standard teacher--student OPD.}
We formulate teacher-supervision allocation as student-side routing before teacher inference, deciding which prompts and rollouts receive teacher supervision without modifying the underlying OPD procedure.

\item \textbf{Success-referenced rollout selection.}
We introduce success-referenced rollout selection, using a same-prompt successful rollout as a teacher-free reference for selecting among failed rollouts. PDA measures sustained representational divergence from this reference, while estimated teacher-input cost accounts for query cost.

\item \textbf{Strong performance--cost efficiency.}
Across three teacher--student pairs and six mathematical reasoning benchmarks, SR-OPD uses only 3.46--5.02\% of Vanilla OPD's teacher-input tokens in the one-pass setting while maintaining comparable reasoning performance. Under a matched 5\% teacher-input budget, controlled experiments support both prompt routing based on successful and failed rollouts and success-referenced rollout selection.

\end{enumerate}

%% file: sections/02_related.tex
\section{Related Work}
\label{sec:related}

\paragraph{Supervision signals in on-policy distillation.}
Recent work has explored how to identify where teacher supervision is most useful. TA-OPD~\citep{wang2026taopd} focuses on the teachability of token-level teacher signals, while PG-OPD~\citep{zhang2026fullrollouts} uses teacher--student compatibility over short prefixes to estimate rollout value. These studies show that teacher or teacher--student information can guide supervision allocation. However, obtaining such information may itself require teacher computation. This raises a complementary question: can supervision value be estimated using only information already available from the student's own rollouts?

\paragraph{Same-prompt outcome signals.}
When multiple rollouts are generated for the same prompt, their outcomes provide direct student-side information about what the student can solve and where it still fails. Recent distillation studies provide evidence for the value of such signals. ReNIO~\citep{lin2026renio} shows that incorrect rollouts can provide stronger learning signals than correct-only rollouts, while MOPD~\citep{yu2026mopd} shows that successful and failed same-prompt rollouts provide complementary information for teacher supervision. Together, these findings suggest that both failure information and the contrast between successful and failed behavior can help identify where supervision is valuable. However, outcome labels remain coarse: multiple failed rollouts from the same prompt share the same failure label even though their reasoning processes may differ substantially.

\paragraph{Hidden-state signals for failure selection.}
This limitation motivates finer-grained signals for distinguishing among failed rollouts. Hidden-state trajectories provide such information by revealing how internal representations evolve throughout reasoning. \citet{sun2026trajectories}, \citet{liang2026reasoning}, and \citet{chen2026shear} identify representational differences between correct and incorrect reasoning and use hidden states for verification, candidate selection, or fine-grained credit assignment. These findings motivate using representation geometry as a teacher-free signal for distinguishing among failed rollouts beyond what outcome labels alone can provide.

%% file: sections/03_method.tex
\section{Method}
\label{sec:method}

Motivated by evidence that successful and failed rollouts from the same prompt provide complementary information for teacher supervision, we ask whether the student's own successful behavior can be used before teacher inference to decide where expensive supervision should be allocated. In SR-OPD, successful rollouts are not sent to the teacher; instead, they serve as teacher-free references for routing. This perspective focuses attention on mixed-outcome prompts: their failed rollouts provide candidates for supervision, while their successful rollouts provide same-prompt references for deciding which failure to query.

Based on this principle, SR-OPD introduces a two-stage pre-query routing procedure before standard OPD supervision. For each training prompt, the student first generates multiple rollouts. The first stage, prompt routing, determines whether the prompt should receive a teacher query. For each retained prompt, the second stage, success-referenced rollout routing, compares failed rollouts against same-prompt successful behavior and selects one failure using persistent representational departure together with estimated teacher-input cost. Both routing decisions rely only on student-side information and are made before teacher inference. The selected failed rollout then receives the unchanged standard OPD supervision target. We first define the routing setup in Section~\ref{sec:preliminaries}, then present prompt routing in Section~\ref{sec:prompt-routing} and rollout routing in Section~\ref{sec:success-referenced-scoring}.

\subsection{Routing Setup}
\label{sec:preliminaries}

We first formalize the student-side information available before teacher inference and the candidate sets used by the two routing stages.

In OPD, we use $x$ to denote a training prompt. For each $x$, the student generates $K$ rollouts,
\begin{equation}
\mathcal{G}(x)=\{y^{(1)},\ldots,y^{(K)}\}.
\label{eq:rollout-set}
\end{equation}
A task verifier assigns each rollout an outcome
\begin{equation}
v(x,y)\in\{0,1,\bot\},
\label{eq:verifier}
\end{equation}
where $1$ denotes success, $0$ denotes failure, and $\bot$ denotes an invalid or unevaluable rollout. Rollouts with $v(x,y)=\bot$ are discarded before routing. The remaining rollouts are divided into successful and failed sets,
\begin{equation}
\mathcal{P}_x=\{y\in\mathcal{G}(x):v(x,y)=1\},
\qquad
\mathcal{N}_x=\{y\in\mathcal{G}(x):v(x,y)=0\}.
\label{eq:outcome-sets}
\end{equation}
Here, $\mathcal{P}_x$ contains the successful rollouts available as same-prompt references, while $\mathcal{N}_x$ contains the failed candidates considered for teacher supervision.

Routing changes only which student rollout receives teacher supervision; the OPD objective applied after selection remains unchanged and is given in Appendix~\ref{app:opd-objective}.

\subsection{Prompt Routing}
\label{sec:prompt-routing}

Given the successful and failed sets $\mathcal{P}_x$ and $\mathcal{N}_x$, the first stage decides whether prompt $x$ is eligible for a teacher query. SR-OPD retains a prompt only when both sets are nonempty:
\begin{equation}
q(x)=
\mathbf{1}
\left[
\mathcal{P}_x\neq\varnothing
\land
\mathcal{N}_x\neq\varnothing
\right].
\label{eq:prompt-routing}
\end{equation}
When $q(x)=0$, no teacher query is issued. An all-success prompt contains no failed candidate to supervise, whereas an all-failure prompt lacks the same-prompt successful reference required by the second stage. When $q(x)=1$, $\mathcal{N}_x$ provides the failed candidates and $\mathcal{P}_x$ provides the successful references generated by the current student, so the prompt proceeds to rollout routing.

This gate does not assume that skipped prompts have no learning value; it only restricts teacher queries to prompts for which the proposed second-stage routing rule is defined. For a retained prompt, the remaining question is which failed candidate in $\mathcal{N}_x$ should receive the teacher query.

\subsection{Success-Referenced Rollout Routing}
\label{sec:success-referenced-scoring}

For a routed prompt, all candidates in $\mathcal{N}_x$ share the same failure outcome, so the binary verifier label cannot distinguish which failure should receive teacher supervision. We therefore use successful rollouts in $\mathcal{P}_x$ as student-side references and compare each failed rollout with successful behavior already produced for the same prompt. This provides a finer-grained teacher-free signal for distinguishing among failures. The selector is constructed in two steps: we first quantify persistent representational departure from successful references, and then combine this signal with estimated teacher-input cost.

\paragraph{Hidden-state trajectory alignment.}
This comparison must handle rollouts of different lengths while avoiding the cost of comparing every token-level hidden state. We therefore approximate each full trajectory using $M$ positions on a normalized progress axis, providing a common alignment across rollouts while keeping routing computation bounded. Specifically,
\begin{equation}
t_j=\frac{j-1}{M-1},
\qquad
\ell_{r,j}=
\operatorname{round}\!\left(t_j(L_r-1)\right),
\qquad
j=1,\ldots,M,
\label{eq:trajectory-grid}
\end{equation}
where $t_j\in[0,1]$ denotes relative progress and $L_r$ is the number of response tokens in rollout $r$.

Let $h_{r,\ell_{r,j}}$ denote the student hidden state at the mapped position. We center it using a fixed mean $\mu$:
\begin{equation}
z_{r,j}=h_{r,\ell_{r,j}}-\mu.
\label{eq:centered-state}
\end{equation}
The hidden-state layer, estimation of $\mu$, and related implementation details are provided in Appendix~\ref{app:routing-algorithm}.

For a failed rollout $n\in\mathcal{N}_x$ and a successful reference $p\in\mathcal{P}_x$, we define their representational alignment at position $j$ as
\begin{equation}
s_j^{(n,p)}
=
\frac{1+\cos\!\left(z_{n,j},z_{p,j}\right)}{2}
\in[0,1].
\label{eq:trajectory-alignment}
\end{equation}
Larger values indicate stronger representational alignment.

\paragraph{Persistent-departure area.}
Pointwise similarity describes local alignment but does not capture departure over the full trajectory or distinguish a deviation that later recovers from one that persists. We therefore track how much previously attained alignment remains unrecovered. Define
\begin{equation}
m_j^{(n,p)}
=
\max_{1\le i\le j}s_i^{(n,p)},
\qquad
d_j^{(n,p)}
=
m_j^{(n,p)}-s_j^{(n,p)}
\in[0,1].
\label{eq:departure}
\end{equation}
If alignment later recovers, $d_j^{(n,p)}$ returns toward zero; if the departure persists, it remains positive across subsequent positions. Thus, $d_j^{(n,p)}$ measures unrecovered loss of previously attained alignment rather than absolute distance at a single position.

To capture both the magnitude and duration of this departure, we aggregate $d_j^{(n,p)}$ over normalized reasoning progress using the trapezoidal rule:
\begin{equation}
A_{n,p}^{(M)}
=
\frac{1}{M-1}
\sum_{j=1}^{M-1}
\frac{
d_j^{(n,p)}+d_{j+1}^{(n,p)}
}{2}
\in[0,1].
\label{eq:persistent-area-pair}
\end{equation}
A brief deviation that quickly recovers contributes little area, whereas a larger or more sustained departure produces a larger value. Thus, $A_{n,p}^{(M)}$ summarizes persistent representational departure over the reasoning trajectory.

For multiple successful references, we define
\begin{equation}
A_n
=
\max_{p\in\mathcal{P}_x}
A_{n,p}^{(M)}
\in[0,1].
\label{eq:persistent-area}
\end{equation}
This is the \textbf{persistent-departure area (PDA)} introduced earlier. PDA measures sustained representational departure of a failed rollout from same-prompt successful behavior and serves as the student-side routing signal. It does not localize semantic errors or directly estimate the true learning utility of a rollout. Its interpretation is discussed in Appendix~\ref{app:score-properties}; the complete computation procedure and proofs of its mathematical properties are given in Appendix~\ref{app:routing-algorithm}.

\paragraph{Cost-aware rollout selection.}
PDA ranks failed rollouts by persistent representational departure, but rollout selection should also account for teacher-input cost. Failed rollouts can differ substantially in length, so candidates with similar PDA values may require different numbers of teacher-input tokens.

For a failed rollout $n$, we define
\begin{equation}
C_n=L_{\mathrm{prompt}}(x)+L_n,
\qquad
C_{\min}=\min_{n\in\mathcal{N}_x}C_n,
\label{eq:query-cost}
\end{equation}
where $C_n$ is measured in serialized teacher-input tokens. Since $C_{\min}\le C_n$, the relative cost factor satisfies $C_{\min}/C_n\in(0,1]$.

We combine this factor with PDA to define
\begin{equation}
S_n
=
A_n\frac{C_{\min}}{C_n}
\in[0,1],
\qquad
n^\star
=
\operatorname*{arg\,max}_{n\in\mathcal{N}_x}S_n.
\label{eq:routing-score}
\end{equation}
The PDA term favors stronger persistent departure, while the cost factor discounts candidates requiring more teacher-input tokens. SR-OPD therefore prioritizes failed rollouts with stronger persistent representational departure and lower relative teacher-input cost.

Because $C_{\min}$ is constant within a prompt, ranking by $S_n$ is equivalent to ranking by $A_n/C_n$. The normalization does not change candidate ordering but keeps $S_n$ within $[0,1]$, consistent with PDA. Accordingly, $S_n$ is a within-prompt supervision-efficiency score rather than a global budget-allocation objective.

SR-OPD sends only the highest-scoring failed rollout $n^\star$ to the teacher. Successful rollouts remain student-side references and are never included in the teacher context; after routing, $n^\star$ receives the unchanged standard OPD supervision and update. The complete deterministic routing procedure and tie-breaking rules are provided in Appendix~\ref{app:routing-algorithm}.

%% file: sections/07_experiments.tex
\section{Experiments}
\label{sec:experiments}

We organize the experiments as an evidence chain from end-to-end efficiency
to stage-level isolation and mechanism analysis. After describing the common
experimental setup in Section~\ref{sec:setup}, we evaluate the native
performance--cost trade-off in Section~\ref{sec:main-results}: can SR-OPD
reduce teacher-input cost while preserving the reasoning performance of
standard OPD? We then isolate the two routing decisions under a matched
teacher-input budget in Section~\ref{sec:ablation}: does mixed-outcome prompt
routing improve budget allocation, and, conditional on a routed prompt, does
success-referenced rollout selection improve over alternative pre-query
selectors? Finally, Section~\ref{sec:empirical-analysis} examines selector
components and whether teacher-free routing scores align with signals
available after teacher inference.

\subsection{Experimental Setup}
\label{sec:setup}

We evaluate three teacher--student pairs:
Qwen3-4B-Instruct-2507~\citep{qwen2025qwen34b2507} $\rightarrow$
Qwen3-1.7B~\citep{yang2025qwen3} (Qwen),
Skywork-OR1-Math-7B~\citep{he2025skywork} $\rightarrow$
DeepSeek-R1-Distill-Qwen-1.5B~\citep{deepseek2025r1} (Skywork), and
Granite-3.3-8B-Instruct~\citep{ibm2025granite338b} $\rightarrow$
Granite-3.3-2B-Instruct~\citep{ibm2025granite332b}
(Granite). The students are trained on DAPO-Math-17k~\citep{yu2025dapo}
and evaluated on AIME24~\citep{maaamc}, AIME25~\citep{maaamc},
AMC23~\citep{maaamc}, HMMT24~\citep{hmmtarchive},
HMMT25~\citep{hmmtarchive}, and
MATH-500~\citep{lightman2024verify,hendrycks2021math}.
Each prompt produces $K=4$ student rollouts. We use three paired training
seeds (41--43) and 16 generation seeds (42--57) per evaluation problem.
Macro Avg is the unweighted mean over the six benchmarks. Teacher cost is
measured in serialized teacher-input tokens. Additional protocol,
accounting, and statistical details are in
Appendix~\ref{app:experimental-details}. With this common setup, we first
examine the native end-to-end performance--cost trade-off.

\subsection{End-to-End Performance--Cost Trade-off}
\label{sec:main-results}

We begin with the end-to-end question: how much teacher-input cost can
SR-OPD remove while maintaining the reasoning performance of standard OPD?
Table~\ref{tab:end-to-end} compares the native operating points of Vanilla
OPD, TA-OPD, TLR-OPD, and SR-OPD in reasoning performance and teacher-input
cost. Vanilla OPD queries the teacher for every eligible rollout. TA-OPD selects 5\% of
teacher-supervised tokens after teacher inference, whereas TLR-OPD adapts
pre-query rollout routing to teacher--student OPD. Their implementation and
cost accounting are described in Appendix~\ref{app:experimental-details}.

\input{sections/generated_native_results}

SR-OPD uses 3.64\%, 3.46\%, and 5.02\% of Vanilla OPD's teacher-input tokens
on Qwen, Skywork, and Granite, respectively. The corresponding paired
Macro Avg differences are $+0.063$, $-0.089$, and $-0.075$ points;
all three 95\% intervals include zero. TA-OPD remains close to Vanilla OPD
in teacher-input use because its selection follows teacher inference.
TLR-OPD reduces teacher input but does not consistently preserve
Vanilla-level performance. Across these native points, SR-OPD has the
lowest observed teacher-input use on all three pairs while maintaining
similar reasoning performance. Appendix~\ref{app:experimental-details}
gives the cost accounting and statistical procedure behind this comparison.
These native comparisons establish the observed end-to-end performance--cost
trade-off. We next move to a matched-budget setting to isolate prompt and
rollout routing from the large differences in native teacher-input use.

\subsection{Controlled Routing Studies}
\label{sec:ablation}

We isolate the two routing stages at the same nominal 5\% Vanilla
teacher-input budget. Eligible queries are subsampled with probability
0.05, and training stops when serialized teacher-input use reaches the
target; additional data passes are allowed. Budget details and realized
accounting are in Appendix~\ref{app:experimental-details}.
We first vary the prompt gate while fixing rollout selection to Random.
We then fix the mixed-outcome gate and vary only the failed-rollout selector.

\begin{figure}[!t]
\centering
\includegraphics[width=\textwidth]{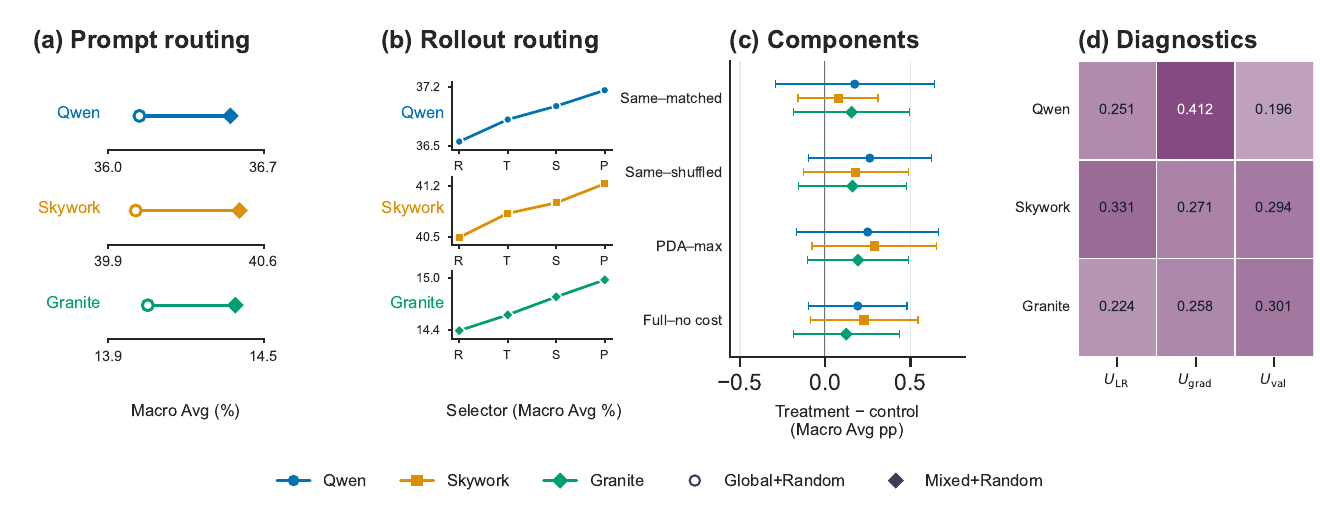}
\caption{Controlled routing and mechanism analyses. (a) Absolute
Macro Avg for Global+Random and Mixed+Random; each model has its own
horizontal score scale. (b) Absolute Macro Avg for the four Mixed-gate
selectors, with a separate local scale per model; R, T, S, and P denote
Random, TLR, Static SR, and PDA. Paired differences and intervals are in
Appendix~\ref{app:rollout-routing-results}.
(c) Independent component effects shown as treatment-minus-control
paired means and pointwise 95\% intervals. (d) Spearman correlations
between the Full SR score and three post-query diagnostics; darker cells
indicate stronger association. Definitions and numerical results are in
Appendix~\ref{app:prompt-gate} for (a),
Appendices~\ref{app:controlled-selectors}--\ref{app:rollout-routing-results}
for (b), Appendix~\ref{app:mechanism-ablations} for (c), and
Appendix~\ref{app:teacher-diagnostics} for (d).}
\label{fig:controlled-routing-evidence}
\end{figure}

\paragraph{Prompt routing.}
To isolate the prompt-level decision, we compare Global+Random, which draws
from the full training data, with Mixed+Random, which retains only prompts
with both valid successful and failed rollouts. Both arms fix rollout
selection to Random.
Figure~\ref{fig:controlled-routing-evidence}(a) shows both absolute
Macro Avg means. The paired Mixed-minus-Global effects are
$+0.410$, $+0.467$, and $+0.338$ Macro Avg points. All three pointwise
95\% intervals lie above zero. Eligibility rules, arm means, and intervals
are in Appendix~\ref{app:prompt-gate} and
Table~\ref{tab:prompt-gate-effects}.
We therefore use the Mixed gate for the subsequent selector comparison.

\paragraph{Rollout routing.}
Having isolated the prompt-level effect, we now fix the mixed-outcome gate
and vary only the failed-rollout selector. Mixed+Random serves as the
prompt-only control.
Figure~\ref{fig:controlled-routing-evidence}(b) shows absolute Macro Avg
for Random, TLR, Static SR, and PDA under the Mixed gate. Relative to
Mixed+Random, the paired SR-OPD gains are
$+0.616$, $+0.738$, and $+0.587$ points on Qwen, Skywork, and Granite;
the Qwen and Skywork intervals exclude zero. The observed differences
relative to TLR and Static SR are also positive, although most intervals
include zero. In particular, all three SR-OPD-minus-Static SR intervals
include zero. The evidence for success-referenced selection over
Mixed+Random is therefore stronger than the evidence for the incremental
effect of persistent departure over Static SR with three training seeds.
Selector definitions are in
Appendix~\ref{app:controlled-selectors}; benchmark means and paired
contrasts are in Appendix~\ref{app:rollout-routing-results} and
Tables~\ref{tab:rollout-routing}--\ref{tab:selector-contrasts}.
Together, the two controlled comparisons support the hierarchical design of
SR-OPD: prompt routing first determines where the teacher budget is spent,
and rollout routing then determines which failure within a selected prompt
receives that supervision.

\subsection{Mechanism Analysis}
\label{sec:empirical-analysis}

The controlled studies show the observed stage-level differences, but do not
identify which elements of the success-referenced selector drive its behavior
or whether its teacher-free score aligns with signals available after teacher
inference. We therefore examine the mechanism at two levels: component
ablations test the reference, trajectory, and cost terms individually, while
teacher-side diagnostics compare pre-query scores with post-query proxies.

\paragraph{Routing components.}
We separately ablate the reference choice, the trajectory-departure
formulation, and the teacher-input cost term.
Figure~\ref{fig:controlled-routing-evidence}(c) compares same-prompt
references with matched and shuffled other-prompt references, PDA with
maximum departure, and the full cost-aware score with its no-cost variant.
PDA exceeds maximum departure by $+0.249$, $+0.287$, and $+0.192$ points;
the full score exceeds its no-cost variant by $+0.191$, $+0.228$, and
$+0.123$ points. All twelve component means are positive, but their
pointwise 95\% intervals include zero. We therefore regard them as
directional evidence. Definitions and paired results are in
Appendix~\ref{app:mechanism-ablations} and
Table~\ref{tab:mechanism-ablations-app}.

\paragraph{Teacher-side diagnostics.}
The component ablations evaluate downstream performance, but they do not
directly test whether teacher-free routing scores rank failures consistently
with signals available only after teacher inference. We compare scores over
200 diagnostic failures per teacher--student pair.
The heatmap in Figure~\ref{fig:controlled-routing-evidence}(d) compares the pre-query
Full SR score with three post-query teacher-side diagnostics. Its Spearman
correlation with local update magnitude, $U_{\mathrm{grad}}$, is 0.412,
0.271, and 0.258 on Qwen, Skywork, and Granite. These associations show
agreement with local post-query signals, without establishing long-horizon
learning utility. Diagnostic definitions and the full correlation matrix
are in Appendix~\ref{app:teacher-diagnostics} and
Table~\ref{tab:utility-ladder}.

%% file: sections/generated_native_results.tex
\begin{table}[H]
\caption{Native operating points. Benchmark columns report accuracy over 16 generation seeds, averaged across three training seeds. Macro Avg is the unweighted mean across six benchmarks for each training seed, then averaged across training seeds. Input is mean serialized teacher-input tokens in millions; parenthesized ratios are decimal values relative to Vanilla OPD in the same pair. Bold marks the highest observed benchmark and Macro Avg mean within each pair and the lowest teacher input.}
\label{tab:end-to-end}
\centering\fontsize{8}{9}\selectfont\apptablestyle
\begin{tabular}{llrrrrrrrr}
\toprule
\multicolumn{1}{c}{Pair} & \multicolumn{1}{c}{Method} & \multicolumn{1}{c}{AIME24} & \multicolumn{1}{c}{AIME25} & \multicolumn{1}{c}{AMC23} & \multicolumn{1}{c}{HMMT24} & \multicolumn{1}{c}{HMMT25} & \multicolumn{1}{c}{MATH-500} & \multicolumn{1}{c}{Macro Avg} & \multicolumn{1}{c}{Input (M; ratio)}\\
\midrule
\multirow{4}{*}{\textbf{Qwen}} & Vanilla OPD & \textbf{34.31} & 19.44 & 61.30 & 11.60 & \textbf{13.96} & \textbf{84.30} & 37.48 & 374.10 (1.00)\\
 & TA-OPD & 29.72 & \textbf{21.39} & 61.35 & \textbf{14.31} & 13.75 & 83.49 & 37.34 & 417.90 (1.12)\\
 & TLR-OPD & 17.57 & 15.35 & 46.98 & 14.10 & 12.08 & 75.57 & 30.27 & 19.20 (0.05)\\
 & \SROPD & 34.17 & 19.58 & \textbf{61.56} & 12.08 & 13.68 & 84.21 & \textbf{37.55} & \textbf{13.60 (0.04)}\\
\midrule
\multirow{4}{*}{\textbf{Skywork}} & Vanilla OPD & \textbf{36.74} & 26.67 & 67.60 & 15.14 & \textbf{19.24} & 85.89 & \textbf{41.88} & 519.80 (1.00)\\
 & TA-OPD & 35.07 & \textbf{26.81} & 66.72 & 14.93 & 17.22 & 85.73 & 41.08 & 516.20 (0.99)\\
 & TLR-OPD & 34.79 & 26.46 & 65.57 & \textbf{15.49} & 17.85 & 84.76 & 40.82 & 126.70 (0.24)\\
 & \SROPD & 36.32 & 26.25 & \textbf{67.71} & 15.14 & 19.17 & \textbf{86.15} & 41.79 & \textbf{18.00 (0.03)}\\
\midrule
\multirow{4}{*}{\textbf{Granite}} & Vanilla OPD & 3.75 & 1.46 & \textbf{25.21} & 3.13 & 0.56 & \textbf{56.99} & \textbf{15.18} & 101.30 (1.00)\\
 & TA-OPD & 3.06 & 1.60 & 23.44 & 2.43 & 0.63 & 55.49 & 14.44 & 95.00 (0.94)\\
 & TLR-OPD & \textbf{4.51} & \textbf{2.78} & 19.90 & \textbf{4.10} & \textbf{2.99} & 40.00 & 12.38 & 5.55 (0.05)\\
 & \SROPD & 3.68 & 1.67 & 25.05 & 2.92 & 0.42 & 56.90 & 15.11 & \textbf{5.09 (0.05)}\\
\bottomrule
\end{tabular}
\end{table}

%% file: sections/08_conclusion.tex
\section{Discussion and Limitations}
\label{sec:discussion}

\paragraph{Discussion.}
The results suggest that teacher-supervision allocation can be decomposed
into two distinct decisions: which prompts should trigger a teacher query,
and which failed rollout within those prompts should receive supervision.
This hierarchical view reframes teacher supervision from uniform processing
of student rollouts into a pre-query allocation problem, and motivates the
separate controlled analyses of prompt routing and rollout routing. The
stage-level experiments provide the clearest evidence for these two routing
decisions, while the component ablations and teacher-side diagnostics are
better viewed as complementary mechanism evidence rather than as
independently resolved performance gains.

This framework also suggests two natural directions for future work. First,
this study focuses primarily on pre-query rollout selection, which reduces
the number of rollouts sent to the teacher before teacher inference. A
complementary direction is post-query token selection, which determines how
to use teacher information after it has already been obtained. Future work
could study their joint optimization, with the goal of reducing the number
of teacher queries while also improving the utilization of the supervision
returned by each query, thereby further improving the performance--efficiency
trade-off of OPD.

Second, the rollout-selection mechanism studied here is based on comparing
multiple student-generated rollouts for the same prompt. This idea is not
inherently restricted to teacher--student OPD. An important extension is to
examine whether success-referenced selection can be incorporated into
multi-rollout training paradigms such as GRPO, where it may serve as a more
general trajectory-selection mechanism for identifying training rollouts
that are more useful among multiple candidates.

\paragraph{Limitations.}
Teacher cost counts serialized teacher-input tokens, excluding student
generation, verification, hidden-state routing, and optimization; the savings
do not establish equivalent reductions in total compute or wall-clock time.
The controlled studies match teacher-input use, not prompt exposure or
optimizer updates. With three paired training seeds, component effects remain
directional rather than independently resolved.

SR-OPD requires reliable verifier outcomes and mixed-outcome prompts; its
success reference is unavailable for all-success or all-failure prompts.
Evaluation covers mathematical reasoning, three teacher--student pairs, and
a limited range of rollout counts, so generalization remains open.
Rollout-count sensitivity and verifier robustness are reported in
Appendices~\ref{app:rollout-count-sensitivity}
and~\ref{app:verifier-robustness}.

\section{Conclusion}
\label{sec:conclusion}

We introduced Success-Referenced On-Policy Distillation (SR-OPD), a two-stage
pre-query routing method for standard teacher--student OPD. SR-OPD reduces
unnecessary teacher queries by making both prompt-level and rollout-level
routing decisions before teacher inference. Its central idea is to use
successful behavior already produced by the student as a reference for
allocating teacher supervision, while leaving the standard OPD supervision
objective unchanged.

Across three teacher--student pairs and six mathematical reasoning
benchmarks, SR-OPD uses only 3.46--5.02\% of the teacher-input tokens
required by Vanilla OPD in the one-pass setting while maintaining comparable
reasoning performance. Controlled experiments under a matched teacher-input
budget further support both prompt routing and success-referenced rollout
selection.

These results show that student-generated success can serve not only as an
outcome of training, but also as an internal reference signal for allocating
teacher supervision. More broadly, they suggest that expensive teacher
computation in OPD need not be distributed uniformly across student-generated
rollouts, but can instead be routed using student-side information available
before teacher inference.

%% file: sections/09_statements.tex
\subsection*{AI Use Statement}

Generative AI tools were used for literature retrieval, analysis-code
development, and language editing. All experimental results were obtained
from saved experimental outputs. The authors remain fully responsible for
the method design, experimental analysis, result verification, and final
manuscript content.

\subsection*{Ethics Statement}

This work uses publicly available mathematical-reasoning datasets and
publicly available models, and involves no human subjects or personally
identifiable information. Because SR-OPD relies on verifier outcomes for
routing, systematic verifier errors may introduce selection bias and affect
which rollouts receive teacher supervision.
Appendix~\ref{app:verifier-robustness} reports verifier audits and
label-corruption robustness analyses to assess this potential risk.

\subsection*{Reproducibility Statement}

Sections~\ref{sec:method} and~\ref{sec:experiments} specify the method,
primary experimental settings, random seeds, and teacher-input accounting.
The appendix provides the complete routing procedure, implementation
configuration, statistical methodology, controlled experiments, mechanism
ablations, and robustness analyses. The supplementary material provides the
main experimental outputs, seed-level results, and analysis scripts used to
reproduce the primary tables and figures.

%% file: sections/10_appendix.tex
\section{Supporting Details, Results, and Analysis}

This appendix follows the organization of the main paper. We first provide
the OPD objective, routing-score interpretation, and complete computation
procedure with proofs and implementation details
supporting Section~\ref{sec:method}. We then give the full experimental
protocol and accounting details, followed by the controlled prompt- and
rollout-routing results corresponding to Section~\ref{sec:ablation}.
Finally, we report the mechanism analyses and robustness checks discussed in
Sections~\ref{sec:empirical-analysis} and~\ref{sec:discussion}.

\input{sections/generated_appendix_results}
\input{sections/controlled_additional_results}

\subsection{Sampled-Token OPD Objective}
\label{app:opd-objective}

SR-OPD changes only which rollout receives teacher supervision; the
sampled-token OPD target and update remain unchanged. To make this separation
explicit, we first state the OPD objective applied after routing. In general,
OPD samples student rollouts and uses teacher token supervision on the
student-visited decoding contexts; the objective below gives the sampled-token
implementation used in our experiments.
Let $\pi_{\theta_0}$ be the rollout-generating student, $\pi_{\theta}$ the trainable student, and $\pi_T$ the frozen teacher.
For token $y_{n,t}$ in rollout $n$, let
\[
c_{n,t}
=
(x_n, y_{n,<t})
\]
denote its decoding context.
We define the token-level teacher--student log-probability difference as
\begin{equation}
\widehat{r}_{n,t}
=
\log \pi_{\theta_0}
\left(
y_{n,t}\mid c_{n,t}
\right)
-
\log \pi_T
\left(
y_{n,t}\mid c_{n,t}
\right).
\label{eq:token-target}
\end{equation}

Let $M_{n,t}\in\{0,1\}$ denote the response-token mask.
It retains valid response tokens, including the terminal EOS token, and
excludes prompt tokens, padding, and tokens from invalid or unevaluable
rollouts.
Using $\operatorname{sg}[\cdot]$ to denote stop-gradient, we define the
detached token advantage as
\begin{equation}
a_{n,t}
=
-
\operatorname{sg}
\left[
\widehat{r}_{n,t}
\right].
\label{eq:token-advantage}
\end{equation}

The one-epoch on-policy surrogate is then
\begin{equation}
\mathcal{L}_{\mathrm{actor}}(\theta)
=
-
\frac{1}{
\sum_{n,t} M_{n,t}
}
\sum_{n,t}
M_{n,t}
a_{n,t}
\frac{
\pi_{\theta}
\left(
y_{n,t}\mid c_{n,t}
\right)
}{
\pi_{\theta_0}
\left(
y_{n,t}\mid c_{n,t}
\right)
}.
\label{eq:opd-objective}
\end{equation}

\subsection{Routing-Score Interpretation}
\label{app:score-properties}

PDA records the magnitude and duration of unrecovered alignment loss, rather
than a token-level semantic error location. Its cost-normalized score ranks
failed rollouts within one prompt. Neither quantity directly measures true
learning utility, and the ranking does not imply optimal allocation across
prompts. Appendix~\ref{app:routing-algorithm} gives the complete calculation
and proves the score properties used in the main text.

\subsection{Complete Computation, Implementation, and Proofs}
\label{app:routing-algorithm}

Here we give the exact trajectory-alignment implementation, complete
single-query procedure, and proofs of the stated score properties.

\begin{algorithm}[t]
\caption{SR-OPD Pre-Query Routing}
\label{alg:sr-opd-routing}
\begin{algorithmic}[1]
\Require prompt $x$; student rollouts $\mathcal{G}(x)$; verifier $v$; trajectory resolution $M$
\State Discard rollouts with $v(x,y)=\bot$
\State Partition valid rollouts into $\mathcal{P}_x$ and $\mathcal{N}_x$ using Eq.~\eqref{eq:outcome-sets}
\If{$\mathcal{P}_x=\varnothing$ \textbf{or} $\mathcal{N}_x=\varnothing$}
  \State \Return no teacher query
\EndIf
\State $C_{\min}\gets\min_{n\in\mathcal{N}_x} C_n$
\For{each failed rollout $n\in\mathcal{N}_x$}
  \For{each success reference $p\in\mathcal{P}_x$}
    \State Compute trajectory alignment $\{s_j^{(n,p)}\}_{j=1}^{M}$ using Eq.~\eqref{eq:trajectory-alignment}
    \State Compute persistent departure $\{d_j^{(n,p)}\}_{j=1}^{M}$ using Eq.~\eqref{eq:departure}
    \State Compute $A_{n,p}^{(M)}$ using Eq.~\eqref{eq:persistent-area-pair}
  \EndFor
  \State $A_n\gets\max_{p\in\mathcal{P}_x}A_{n,p}^{(M)}$
  \State $S_n\gets A_n\,C_{\min}/C_n$ using Eq.~\eqref{eq:routing-score}
\EndFor
\State Select $n^\star$ using Eq.~\eqref{eq:routing-score} and the deterministic tie-break below
\State Query the teacher only for $n^\star$
\State Apply the unchanged sampled-token OPD objective from Appendix~\ref{app:opd-objective}
\end{algorithmic}
\end{algorithm}

\paragraph{Representation computation.}
For each rollout, we use student hidden states from the final normalization
layer immediately before the language-model head. For trajectory resolution
$M\ge2$, normalized response positions are
\begin{equation}
t_j=\frac{j-1}{M-1},\qquad j=1,\ldots,M.
\label{eq:normalized-grid}
\end{equation}
For rollout $r$ with $L_r$ response tokens, we map to response-token index
\begin{equation}
\ell_{r,j}=\operatorname{round}\!\left(t_j(L_r-1)\right).
\label{eq:token-alignment}
\end{equation}
At this mapped response-token position, write
$h_{r,j}:=h_{r,\ell_{r,j}}$. The centered state is
\begin{equation}
z_{r,j}=h_{r,j}-\mu,
\label{eq:centered-state-implementation}
\end{equation}
where $\mu$ is estimated once from held-out student states and held fixed.
We compute cosine similarities in FP32 and accumulate PDA at no lower than
FP32 precision. For responses shorter than $M$, several grid positions may
map to one token; they remain distinct grid locations, and all $M-1$
intervals contribute to Eq.~\eqref{eq:persistent-area-pair}.
Empty or unevaluable generations are removed before prompt routing.
Additional configuration and accounting details appear in
Appendix~\ref{app:full-config}.

\paragraph{Determinism.}\label{app:tie-break}
Ties are resolved by larger PDA, lower estimated cost, and smaller rollout ID, in that order. If all PDA values are zero, the rule reduces to least-cost selection. The procedure is deterministic given the rollouts, verifier outcomes, and hidden states.

\paragraph{Theoretical properties and proofs.}
Let $\widetilde d_M(t)$ be the piecewise-linear interpolation of
$d_1^{(n,p)},\ldots,d_M^{(n,p)}$ on the uniform grid in
Eq.~\eqref{eq:normalized-grid}. The trapezoidal rule is exact for this
interpolant, so
\begin{equation}
A_{n,p}^{(M)}=\int_0^1\widetilde d_M(t)\,dt.
\label{eq:pda-integral}
\end{equation}
Because cosine similarity lies in $[-1,1]$, Eq.~\eqref{eq:trajectory-alignment}
gives $0\le s_j^{(n,p)}\le1$. Hence
$0\le d_j^{(n,p)}=m_j^{(n,p)}-s_j^{(n,p)}\le1$, and the trapezoidal
average and maximum over references give $0\le A_{n,p}^{(M)},A_n\le1$.
The nonnegative interpolant has zero integral exactly when every grid
departure is zero, which is equivalent to the alignment sequence never
falling below its running maximum. If $\widetilde d_M(t)\ge\Delta$ on a
measurable set $E\subseteq[0,1]$, nonnegativity elsewhere yields
$A_{n,p}^{(M)}\ge\Delta\,\operatorname{Leb}(E)$; thus a departure's
duration as well as its magnitude contributes to PDA.

For $C_n\ge C_{\min}>0$, the factor $C_{\min}/C_n$ lies in $(0,1]$,
so $0\le S_n\le1$. Because $C_{\min}$ is positive and constant for all
failed candidates of the same prompt,
\begin{equation}
\operatorname*{arg\,max}_{n\in\mathcal{N}_x} A_n\frac{C_{\min}}{C_n}
=\operatorname*{arg\,max}_{n\in\mathcal{N}_x}\frac{A_n}{C_n}.
\label{eq:value-density-equivalence}
\end{equation}
This proves within-prompt ranking equivalence; it makes no claim about a
global budget optimum or a candidate's actual learning utility.

\input{sections/appendix_experiments_A4_A9}

%% file: sections/generated_appendix_results.tex
\newcommand{\RolloutResults}{%
\subsection{Controlled Rollout Routing}\label{app:rollout-routing-results}\label{app:full-results}
The teacher-budget-matched controlled comparison fixes the mixed-outcome
prompt gate at the 5\% teacher-input budget and contrasts five failed-rollout
selectors. We report paired selector effects below.

\input{sections/generated_selector_contrasts}

}

\newcommand{\RolloutCountResults}{%
\subsection{Rollout-Count Sensitivity}\label{app:rollout-count-sensitivity}
This is a separate 3\% cap sensitivity run comparing $K=8$ and $K=16$.
It does not include the primary $K=4$, 5\% configuration.
Table~\ref{tab:k-teacher-input} reports realized teacher-input usage under
the same serialized-token accounting definition as the main experiments.
Realized ratios remain close to the nominal 3\% cap across both rollout counts
and all routing methods.
\begin{table}[!htbp]
\caption{Rollout-count sensitivity at the nominal 3\% cap. The final column is the paired $K=16$ minus $K=8$ change.}
\label{tab:k8-e2e}
\centering
\setlength{\tabcolsep}{2pt}
\begin{tabular}{lcccc}
\toprule
\multicolumn{1}{c}{Pair} & \multicolumn{1}{c}{Method} & \multicolumn{1}{c}{$K=8$} & \multicolumn{1}{c}{$K=16$} & \multicolumn{1}{c}{Change [95\% CI]}\\
\midrule
\multirow{5}{*}{Qwen} & Mixed+Random & 36.49 & 36.53 & $+0.040\;[-0.072,+0.152]$\\
  & Cost-only & 36.85 & 36.90 & $+0.050\;[-0.049,+0.149]$\\
  & Frontier-TLR & 36.79 & 36.82 & $+0.030\;[-0.082,+0.142]$\\
  & Static SR & 36.99 & 37.05 & $+0.060\;[-0.052,+0.172]$\\
  & SR-OPD & 37.16 & 37.32 & $+0.160\;[+0.036,+0.284]$\\
\multirow{5}{*}{Skywork} & Mixed+Random & 40.39 & 40.48 & $+0.090\;[-0.022,+0.202]$\\
  & Cost-only & 40.69 & 40.73 & $+0.040\;[-0.059,+0.139]$\\
  & Frontier-TLR & 40.69 & 40.70 & $+0.010\;[-0.102,+0.122]$\\
  & Static SR & 40.94 & 40.88 & $-0.060\;[-0.184,+0.064]$\\
  & SR-OPD & 41.18 & 41.28 & $+0.100\;[-0.024,+0.224]$\\
\multirow{5}{*}{Granite} & Mixed+Random & 14.30 & 14.33 & $+0.030\;[-0.069,+0.129]$\\
  & Cost-only & 14.60 & 14.67 & $+0.070\;[-0.029,+0.169]$\\
  & Frontier-TLR & 14.48 & 14.46 & $-0.020\;[-0.107,+0.067]$\\
  & Static SR & 14.65 & 14.72 & $+0.070\;[-0.029,+0.169]$\\
  & SR-OPD & 14.90 & 14.97 & $+0.070\;[-0.029,+0.169]$\\
\bottomrule
\end{tabular}
\end{table}

\begin{table}[!ht]
\caption{Realized teacher-input usage for the rollout-count sensitivity
experiment. Input is serialized teacher-input tokens in millions; percentages
are relative to the corresponding Vanilla OPD teacher-input mean.}
\label{tab:k-teacher-input}
\centering
\fontsize{8}{9}\selectfont
\setlength{\tabcolsep}{2pt}
\begin{tabular}{llrrrr}
\toprule
Pair & Method & $K=8$ input (M) & $K=8$ / Vanilla (\%) & $K=16$ input (M) & $K=16$ / Vanilla (\%)\\
\midrule
\multirow{5}{*}{Qwen} & Mixed+Random & 11.350 & 3.034 & 11.375 & 3.041\\
  & Cost-only & 11.130 & 2.975 & 11.114 & 2.971\\
  & Frontier-TLR & 11.306 & 3.022 & 11.333 & 3.030\\
  & Static SR & 11.207 & 2.996 & 11.242 & 3.005\\
  & SR-OPD & 11.207 & 2.996 & 11.215 & 2.998\\
\multirow{5}{*}{Skywork} & Mixed+Random & 15.551 & 2.992 & 15.590 & 2.999\\
  & Cost-only & 15.448 & 2.972 & 15.427 & 2.968\\
  & Frontier-TLR & 15.627 & 3.006 & 15.651 & 3.011\\
  & Static SR & 15.546 & 2.991 & 15.560 & 2.993\\
  & SR-OPD & 15.663 & 3.013 & 15.673 & 3.015\\
\multirow{5}{*}{Granite} & Mixed+Random & 3.038 & 2.999 & 3.042 & 3.003\\
  & Cost-only & 3.017 & 2.978 & 3.017 & 2.978\\
  & Frontier-TLR & 3.042 & 3.003 & 3.043 & 3.004\\
  & Static SR & 3.032 & 2.993 & 3.031 & 2.992\\
  & SR-OPD & 3.038 & 2.999 & 3.040 & 3.001\\
\bottomrule
\end{tabular}
\end{table}

}

\newcommand{\MechanismResults}{%
\subsection{Mechanism Ablations}\label{app:mechanism-ablations}\label{app:reference-ablation}\label{app:component-ablation}
\paragraph{Component ablations.}
The four independent contrasts test same-prompt versus matched and shuffled success references, persistent area versus maximum departure, and the contribution of cost normalization. These four comparisons are the complete set of component ablations conducted for the components reported here; no completed comparison was excluded based on the direction or magnitude of its effect. Each row compares its own paired treatment and control runs.
\begin{table}[!htbp]
\caption{Independent paired ablation reruns at the matched 5\% teacher-input
budget. Means and pointwise 95\% intervals use paired seeds 41--43. Each
contrast uses its own control and treatment runs; treatment absolute means
are not shared across rows.}
\label{tab:mechanism-ablations-app}
\centering
\setlength{\tabcolsep}{2pt}
\begin{tabular}{lcc}
\toprule
\multicolumn{1}{c}{Comparison} & \multicolumn{1}{c}{Pair} & \multicolumn{1}{c}{Difference [95\% CI]}\\
\midrule
\multirow{3}{*}{Same $-$ matched} & Qwen & $+0.173\;[-0.293,+0.639]$\\
  & Skywork & $+0.076\;[-0.159,+0.310]$\\
  & Granite & $+0.154\;[-0.184,+0.493]$\\
\multirow{3}{*}{Same $-$ shuffled} & Qwen & $+0.262\;[-0.098,+0.621]$\\
  & Skywork & $+0.180\;[-0.129,+0.488]$\\
  & Granite & $+0.160\;[-0.156,+0.477]$\\
\multirow{3}{*}{PDA $-$ max departure} & Qwen & $+0.249\;[-0.168,+0.666]$\\
  & Skywork & $+0.287\;[-0.077,+0.651]$\\
  & Granite & $+0.192\;[-0.104,+0.488]$\\
\multirow{3}{*}{Full $-$ no cost} & Qwen & $+0.191\;[-0.098,+0.479]$\\
  & Skywork & $+0.228\;[-0.088,+0.544]$\\
  & Granite & $+0.123\;[-0.187,+0.434]$\\
\bottomrule
\end{tabular}
\end{table}

}

\newcommand{\DiagnosticResults}{%
\subsection{Teacher-Side Diagnostics}\label{app:teacher-diagnostics}
These Spearman correlations measure rank association between student-side
routing scores and three post-query diagnostic proxies on the diagnostic set.
For each pair, we fix the training-initial student checkpoint $\theta_0$ and
the same frozen teacher used in the main experiments. Using seed 42, we
randomly order prompts from the DAPO-Math-17k training set, generate four
student rollouts per prompt with the training decoding settings, and retain
mixed-outcome prompts with at least one valid success and failure. We choose
the lowest-index valid failure from each retained prompt and continue scanning
until 200 diagnostic items have been collected. A separate 200-example
validation subset is sampled from DAPO-Math-17k with seed 43 after excluding
the diagnostic prompt IDs. This subset is disjoint from the diagnostic items,
but is drawn from the training dataset; it is therefore a diagnostic validation
subset rather than an independent generalization test.

For each diagnostic failure $y_n$, the frozen teacher scores the same response
tokens used by the OPD objective (Appendix~\ref{app:opd-objective}). With
$M_{n,t}$ denoting its valid response-token mask, the first post-query proxy is
\begin{equation}
U_{\mathrm{LR}}^{(n)}=
\frac{\sum_t M_{n,t}\left[\log\pi_T(y_{n,t}\mid c_{n,t})-
\log\pi_{\theta_0}(y_{n,t}\mid c_{n,t})\right]}
{\sum_t M_{n,t}}.
\end{equation}
A larger $U_{\mathrm{LR}}$ means that the teacher assigns greater likelihood
to the observed failure tokens relative to the student; it is a likelihood
proxy, not a measured learning gain. The second proxy is the gradient norm of
the item-restricted sampled-token OPD loss,
\begin{equation}
U_{\mathrm{grad}}^{(n)}=
\left\|\left.\nabla_\theta\mathcal{L}_{\mathrm{OPD}}^{(n)}(\theta)
\right|_{\theta=\theta_0}\right\|_2.
\end{equation}
A larger norm indicates a stronger local update signal, without establishing
greater downstream utility. For the third proxy, we make one diagnostic
gradient step $\theta_n'=\theta_0-5\times10^{-6}
\left.\nabla_\theta\mathcal{L}_{\mathrm{OPD}}^{(n)}(\theta)
\right|_{\theta=\theta_0}$ and define
\begin{equation}
U_{\mathrm{val}}^{(n)}=
\mathcal{L}_{\mathrm{val}}(\theta_0)-
\mathcal{L}_{\mathrm{val}}(\theta_n'),
\end{equation}
where $\mathcal{L}_{\mathrm{val}}$ is standard-answer token negative
log-likelihood averaged over valid tokens in the fixed validation subset.
Higher $U_{\mathrm{val}}$ means a larger decrease in this diagnostic loss.
Each correlation below is Spearman's $\rho$ across the same 200 failure items
for one pair, comparing an item-level pre-query score with an item-level
post-query proxy. These proxies are never available to the routing policy.
\begin{table}[!htbp]
\caption{Per-pair Spearman correlations between student-side routing scores
and three post-query diagnostic proxies over 200 diagnostic items per
teacher--student pair.}
\label{tab:utility-ladder}
\centering
\setlength{\tabcolsep}{2pt}
\begin{tabular}{lcccc}
\toprule
\multicolumn{1}{c}{Pair} & \multicolumn{1}{c}{Score} & \multicolumn{1}{c}{$U_{\mathrm{LR}}$} & \multicolumn{1}{c}{$U_{\mathrm{grad}}$} & \multicolumn{1}{c}{$U_{\mathrm{val}}$}\\
\midrule
\multirow{4}{*}{Qwen} & TLR & 0.071 & 0.082 & 0.162\\
  & AvgDistance & 0.101 & 0.112 & 0.126\\
  & PDA & 0.271 & 0.301 & 0.314\\
  & FullSR & 0.251 & 0.412 & 0.196\\
\multirow{4}{*}{Skywork} & TLR & 0.104 & 0.143 & 0.151\\
  & AvgDistance & 0.116 & 0.071 & 0.181\\
  & PDA & 0.194 & 0.214 & 0.239\\
  & FullSR & 0.331 & 0.271 & 0.294\\
\multirow{4}{*}{Granite} & TLR & 0.081 & 0.057 & 0.103\\
  & AvgDistance & 0.142 & 0.171 & 0.158\\
  & PDA & 0.281 & 0.244 & 0.371\\
  & FullSR & 0.224 & 0.258 & 0.301\\
\bottomrule
\end{tabular}
\end{table}
}

\newcommand{\VerifierAuditResults}{%
\subsection{Verifier Audit and Robustness}\label{app:verifier-robustness}\label{app:robustness}\label{app:verifier-stress}
\paragraph{Verifier reliability and label corruption.}
We evaluate verifier reliability with a 400-item audit for each
teacher--student pair and evaluate routing robustness by injecting 5\% and
10\% random verifier-label corruption. Table~\ref{tab:verifier-audit}
reports the audit error rates with Wilson intervals together with the
corresponding Macro Avg under label corruption.
\begin{table}[H]
\caption{Verifier reliability: audit error among 400 items per pair with 95\%
Wilson intervals, and Macro Avg under random label flips. The $0\%$ column
is the unperturbed controlled SR-OPD baseline.}
\label{tab:verifier-audit}\label{tab:verifier-stress}
\centering\setlength{\tabcolsep}{2pt}
\begin{tabular}{lcccc}
\toprule
Pair & Audit error [95\% CI] & 0\% flip & 5\% flip & 10\% flip\\
\midrule
Qwen & 1.50\% [0.69,3.23] & 37.17 & 37.05 & 36.84\\
Skywork & 2.25\% [1.19,4.22] & 41.23 & 41.07 & 40.83\\
Granite & 1.00\% [0.39,2.54] & 14.98 & 14.87 & 14.72\\
\bottomrule
\end{tabular}
\end{table}
}

%% file: sections/generated_selector_contrasts.tex
\begin{table}[!htbp]
\caption{Paired SR-OPD minus selector effects at the matched 5\% teacher-input budget. Means and pointwise 95\% intervals use paired seeds 41--43 and the $t_2$ formula.}
\label{tab:selector-contrasts}
\centering\small
\begin{tabular}{llc}
\toprule
Pair & Selector & Difference [95\% CI]\\
\midrule
Qwen & Mixed+Random & $+0.616\;[+0.241,+0.991]$\\
Qwen & Cost-only & $+0.337\;[-0.350,+1.024]$\\
Qwen & Frontier-TLR & $+0.352\;[-0.305,+1.008]$\\
Qwen & Static SR & $+0.191\;[-0.514,+0.895]$\\
Skywork & Mixed+Random & $+0.738\;[+0.651,+0.825]$\\
Skywork & Cost-only & $+0.451\;[-0.044,+0.946]$\\
Skywork & Frontier-TLR & $+0.408\;[+0.255,+0.561]$\\
Skywork & Static SR & $+0.262\;[-0.372,+0.895]$\\
Granite & Mixed+Random & $+0.587\;[-0.112,+1.285]$\\
Granite & Cost-only & $+0.347\;[-0.210,+0.904]$\\
Granite & Frontier-TLR & $+0.405\;[-0.130,+0.940]$\\
Granite & Static SR & $+0.196\;[-0.184,+0.576]$\\
\bottomrule
\end{tabular}
\end{table}

%% file: sections/controlled_additional_results.tex
\newcommand{\PromptResults}{%
\subsection{Controlled Prompt Routing}
\label{app:prompt-gate}

\paragraph{Controlled comparison protocol.}
The prompt- and rollout-routing comparisons use the same paired-seed protocol.
For prompt routing, Mixed+Random is the prompt-only control and Mixed+SR is
the formal SR-OPD arm under the mixed-outcome gate; both reuse the corresponding
rollout-routing runs.

\paragraph{Prompt eligibility and selector compatibility.}
Uniform routing requires at least one valid failed rollout; mixed routing
requires both a valid success and a valid failure. On mixed-outcome prompts,
Random selects a failed rollout at random and SR selects a failure using a
same-prompt success reference. On all-failure prompts, Uniform+Random uses
Random and Uniform+SR falls back to Random; neither Mixed arm queries the
teacher. Prompts without a valid failure receive no query under either gate;
mixed routing also skips prompts without a valid success. Invalid rollouts are
discarded before the gate is applied.
Every emitted teacher query counts its actual serialized teacher-input tokens
against the matched 5\% teacher-input budget; scanning, verification, and routing without
a teacher query do not consume that teacher-input budget.
In this prompt-gating comparison, ``SR'' denotes the SR-OPD rollout selector
used in the formal rollout-routing experiment, not the Static SR ablation.
Mixed+Random is the same prompt-only control as the Mixed+Random arm in
Table~\ref{tab:rollout-routing}, and Mixed+SR is the formal SR-OPD arm under
the mixed-outcome gate. These Mixed-arm means therefore come from the same
teacher-budget-matched controlled rollout-routing runs, not separate training runs.

\begin{table}[!htbp]
\caption{Controlled prompt-routing comparison at the matched 5\%
teacher-input budget. Effects are mixed minus uniform Avg@16 with paired
three-seed $t_2$ 95\% intervals. Statistical details are in
Appendix~\ref{app:statistics}.}
\label{tab:prompt-gate-effects}
\centering\small
\begin{tabular}{llccc}
\toprule
Pair & Selector & Uniform Avg@16 & Mixed Avg@16 & Mixed $-$ uniform [95\% CI]\\
\midrule
\multirow{2}{*}{Qwen} & Random & 36.14 & 36.55 & $+0.410\;[+0.213,+0.607]$\\
  & SR & 36.64 & 37.17 & $+0.525\;[+0.296,+0.753]$\\
\multirow{2}{*}{Skywork} & Random & 40.02 & 40.49 & $+0.467\;[+0.265,+0.669]$\\
  & SR & 40.65 & 41.23 & $+0.578\;[+0.372,+0.783]$\\
\multirow{2}{*}{Granite} & Random & 14.05 & 14.39 & $+0.338\;[+0.130,+0.546]$\\
  & SR & 14.57 & 14.98 & $+0.409\;[+0.216,+0.601]$\\
\bottomrule
\end{tabular}
\end{table}
}

\newcommand{\CostResults}{%
\paragraph{Teacher-input accounting.}
\label{app:teacher-cost-accounting}

The controlled rows report realized serialized teacher-input totals for the
formal mixed-gate rollout-routing comparison. Native input means are reported
in Table~\ref{tab:end-to-end}; the controlled rows compare realized input
with the matched 5\% teacher-input budget. Teacher-input accounting excludes student generation,
routing, verification, and optimization. Per-run data passes, prompt visits,
student rollout counts, eligible prompt counts, teacher-query counts, and
optimizer-update counts are not available in the current analysis records;
the table therefore supports teacher-input matching only.

\begin{table}[!ht]
\caption{Realized teacher-input accounting for the controlled mixed-gate
rollout-routing comparison at the matched 5\% teacher-input budget. Input denotes serialized
teacher-input tokens; percentages are computed relative to the corresponding
Vanilla OPD mean.}
\label{tab:controlled-teacher-cost}
\centering\footnotesize\setlength{\tabcolsep}{3pt}\renewcommand{\arraystretch}{0.95}
\begin{tabular}{llrr}
\toprule Pair & Method & Input (M) & Input / Vanilla (\%)\\\midrule
\multirow{5}{*}{Qwen} & Mixed+Random & 18.777 & 5.02\\
  & Cost-only & 18.667 & 4.99\\
  & Frontier-TLR & 18.787 & 5.02\\
  & Static SR & 18.626 & 4.98\\
  & SR-OPD & 18.730 & 5.01\\
\multirow{5}{*}{Skywork} & Mixed+Random & 26.057 & 5.01\\
  & Cost-only & 25.825 & 4.97\\
  & Frontier-TLR & 26.011 & 5.00\\
  & Static SR & 25.892 & 4.98\\
  & SR-OPD & 26.012 & 5.00\\
\multirow{5}{*}{Granite} & Mixed+Random & 5.080 & 5.02\\
  & Cost-only & 5.034 & 4.97\\
  & Frontier-TLR & 5.055 & 4.99\\
  & Static SR & 5.058 & 4.99\\
  & SR-OPD & 5.068 & 5.00\\
\bottomrule
\end{tabular}
\end{table}
}

%% file: sections/appendix_experiments_A4_A9.tex
\subsection{Additional Experimental Details, Accounting, and Statistics}
\label{app:experimental-details}
\label{app:full-config}

\paragraph{Additional configuration.}
Native experiments use one training-data pass. Hidden-state trajectory
alignment uses $M=16$ positions. The six evaluation benchmarks contain
660 problems in total.

\paragraph{Native baselines.}
Vanilla OPD queries the teacher for every eligible rollout under the
standard OPD procedure. TA-OPD retains 5\% of teacher-supervised tokens
after teacher inference on the queried rollout; it does not avoid that
rollout's teacher-input cost. TLR-OPD adapts the original
TLR~\citep{anonymous2026tlr} pre-query selection rule to standard
teacher--student OPD, querying the teacher only after the student-side
selector chooses a rollout. SR-OPD applies the two routing stages from
Section~\ref{sec:method} before the unchanged OPD update.

\paragraph{Teacher-input accounting.}
Each teacher query contributes its exact serialized input length.
Student generation, verification, routing, and optimization are outside
this accounting.
Controlled experiments target 5\% of the corresponding Vanilla OPD
teacher-input tokens. For each method's eligible candidates, queries
are subsampled with probability 0.05. Training stops when accumulated
teacher-input tokens reach the target; if one pass is insufficient,
subsequent passes use the same routing and sampling rule. A final query
can place realized use slightly above the nominal target.
Table~\ref{tab:controlled-teacher-cost} gives the range across the four
mixed-gate selectors used in the main controlled comparison.

\begin{table}[!htbp]
\caption{Realized teacher-input use across the four mixed-gate selectors,
as a percentage of the corresponding Vanilla OPD input.}
\label{tab:controlled-teacher-cost}
\centering\setlength{\tabcolsep}{2pt}
\begin{tabular}{lc}
\toprule Pair & Range (\%)\\ \midrule
Qwen & 4.98--5.02\\
Skywork & 4.98--5.01\\
Granite & 4.99--5.02\\
\bottomrule
\end{tabular}
\end{table}

\paragraph{Statistics and provenance.}
\label{app:statistics}
The training seed is the statistical unit. Treatment and control are
paired within each teacher--student pair and seed. For paired differences
$d_s$, we report $\bar d\pm t_{0.975,2}s_d/\sqrt{3}$; all 95\% intervals
are pointwise, two-sided, and unadjusted for multiplicity. With three
seeds, they are small-sample uncertainty summaries. The 16 generation
seeds provide repeated samples per evaluation problem rather than
independent training replicates. Benchmark means and paired contrasts
come from \texttt{data/full-data.xlsx}; intermediate values are not
rounded before aggregation. Diagnostic correlations in
Appendix~\ref{app:teacher-diagnostics} are descriptive.

\subsection{Controlled Prompt Routing}
\label{app:prompt-gate}

This comparison fixes rollout selection to Random and changes only the
prompt gate. Global+Random allocates the teacher budget over the full
training data; a teacher query requires a valid failed rollout, selected
uniformly at random. Mixed+Random retains only prompts with at least one
valid success and one valid failure, then selects one valid failure
uniformly. Invalid or unevaluable rollouts are removed before routing.
The Global+Random values in Table~\ref{tab:prompt-gate-effects} correspond
to the source-workbook rows identified by the authors for that experiment.

\begin{table}[!htbp]
\caption{Controlled prompt routing at the matched 5\% teacher-input
budget. Effects are paired Mixed+Random minus Global+Random Macro Avg.}
\label{tab:prompt-gate-effects}
\centering\setlength{\tabcolsep}{2pt}
\begin{tabular}{lcccc}
\toprule
Pair & Global+Random & Mixed+Random & Difference & 95\% CI\\
\midrule
Qwen & 36.14 & 36.55 & $+0.410$ & $[+0.213,+0.607]$\\
Skywork & 40.02 & 40.49 & $+0.467$ & $[+0.265,+0.669]$\\
Granite & 14.05 & 14.39 & $+0.338$ & $[+0.130,+0.546]$\\
\bottomrule
\end{tabular}
\end{table}
All three paired means are positive, and all three pointwise intervals
lie above zero.

\subsection{Controlled Rollout Selectors}
\label{app:controlled-selectors}

All selectors use the same Mixed prompt gate, valid failed-candidate set
$\mathcal N_x$, and matched teacher-input budget. They differ only in
their pre-query ranking of failed rollouts.

\paragraph{Random.}
Random selects one valid failed rollout uniformly and uses no other
routing signal.

\paragraph{Adapted TLR.}
\label{app:frontier-tlr}
The teacher-free TLR~\citep{anonymous2026tlr} adaptation uses response
length $\ell_n$ and mean student token entropy $\bar H_n$, normalized
within $\mathcal N_x$:
\begin{equation}
\begin{aligned}
\widehat\ell_n&=
\frac{\ell_n-\min_{m\in\mathcal N_x}\ell_m}
{\max_{m\in\mathcal N_x}\ell_m-\min_{m\in\mathcal N_x}\ell_m+\epsilon},\\
\widehat H_n&=
\frac{\bar H_n-\min_{m\in\mathcal N_x}\bar H_m}
{\max_{m\in\mathcal N_x}\bar H_m-\min_{m\in\mathcal N_x}\bar H_m+\epsilon},
\quad \epsilon=10^{-8}.
\end{aligned}
\end{equation}
It ranks by $S_n^{\mathrm{TLR}}=(1-\widehat\ell_n)(1-\widehat H_n)$.
A sole failed rollout receives score 1. Ties prefer the shorter rollout
and then the smaller rollout ID.

\paragraph{StaticSR.}
\label{app:static-sr}
StaticSR uses the same successful references, student hidden states,
normalized response grid, reference aggregation, and cost factor as
SR-OPD, but substitutes mean representation distance for persistent
departure. With $s_j^{(n,p)}$ from Section~\ref{sec:method},
\begin{equation}
D_n=\max_{p\in\mathcal P_x}\frac1M\sum_{j=1}^{M}(1-s_j^{(n,p)}),
\qquad
S_n^{\mathrm{StaticSR}}=D_n\frac{C_{\min}}{C_n}.
\end{equation}

\paragraph{Mixed+PDA (SR-OPD).}
The full selector uses the PDA score $A_n$ from
Eq.~\eqref{eq:routing-score} with the same cost normalization:
\begin{equation}
S_n^{\mathrm{SR}}=A_n\frac{C_{\min}}{C_n}.
\end{equation}
Thus, Random uses no ranking signal, TLR uses length and entropy,
StaticSR uses a same-prompt successful reference with static distance,
and SR-OPD uses the same reference with persistent departure.

\subsection{Controlled Rollout Routing}
\label{app:rollout-routing-results}
\label{app:full-results}

We fix the Mixed gate and compare the four failed-rollout selectors in
Table~\ref{tab:rollout-routing}. The benchmark values are means across
three training seeds; Macro Avg weights the six benchmarks equally.

\begin{table}[H]
\caption{Controlled rollout-routing results at the matched 5\%
teacher-input budget.}
\label{tab:rollout-routing}
\centering\fontsize{8}{9}\selectfont\setlength{\tabcolsep}{2pt}
\begin{tabular}{llrrrrrrr}
\toprule
Pair & Method & AIME24 & AIME25 & AMC23 & HMMT24 & HMMT25 &
MATH-500 & Macro Avg\\
\midrule
Qwen & Mixed+Random & 30.90 & 21.39 & 58.70 & 12.71 & 14.72 & 80.88 & 36.55\\
 & Mixed+TLR & 30.97 & 21.39 & 59.38 & 13.13 & 15.14 & 80.89 & 36.81\\
 & Mixed+StaticSR & 31.11 & 21.60 & 59.53 & 13.26 & 15.21 & 81.14 & 36.98\\
 & Mixed+PDA (SR-OPD) & 31.11 & 21.74 & 59.95 & 13.61 & 15.49 & 81.10 & \textbf{37.17}\\
\midrule
Skywork & Mixed+Random & 35.49 & 26.04 & 65.47 & 13.75 & 17.15 & 85.05 & 40.49\\
 & Mixed+TLR & 35.83 & 26.46 & 65.63 & 14.17 & 17.71 & 85.13 & 40.82\\
 & Mixed+StaticSR & 36.04 & 26.53 & 66.09 & 14.24 & 17.64 & 85.27 & 40.97\\
 & Mixed+PDA (SR-OPD) & 36.18 & 26.53 & 66.41 & 14.93 & 17.92 & 85.41 & \textbf{41.23}\\
\midrule
Granite & Mixed+Random & 2.85 & 1.32 & 24.22 & 2.29 & 0.42 & 55.25 & 14.39\\
 & Mixed+TLR & 3.06 & 1.46 & 24.53 & 2.43 & 0.56 & 55.40 & 14.57\\
 & Mixed+StaticSR & 3.19 & 1.81 & 24.48 & 2.85 & 0.76 & 55.60 & 14.78\\
 & Mixed+PDA (SR-OPD) & 3.40 & 1.81 & 24.90 & 3.13 & 1.18 & 55.45 & \textbf{14.98}\\
\bottomrule
\end{tabular}
\end{table}

\begin{table}[H]
\caption{Paired Mixed+PDA minus comparator effects on Macro Avg.
Intervals use paired training seeds 41--43.}
\label{tab:selector-contrasts}
\centering\setlength{\tabcolsep}{2pt}
\begin{tabular}{llcc}
\toprule Pair & Comparator & Difference & 95\% CI\\ \midrule
Qwen & Mixed+Random & $+0.616$ & $[+0.241,+0.991]$\\
 & Mixed+TLR & $+0.352$ & $[-0.305,+1.008]$\\
 & Mixed+StaticSR & $+0.191$ & $[-0.514,+0.895]$\\
Skywork & Mixed+Random & $+0.738$ & $[+0.651,+0.825]$\\
 & Mixed+TLR & $+0.408$ & $[+0.255,+0.561]$\\
 & Mixed+StaticSR & $+0.262$ & $[-0.372,+0.895]$\\
Granite & Mixed+Random & $+0.587$ & $[-0.112,+1.285]$\\
 & Mixed+TLR & $+0.405$ & $[-0.130,+0.940]$\\
 & Mixed+StaticSR & $+0.196$ & $[-0.184,+0.576]$\\
\bottomrule
\end{tabular}
\end{table}
The stage-level comparison with Mixed+Random is stronger than the
incremental comparison with StaticSR. All three StaticSR differences are
positive, but their intervals include zero.

\subsection{Mechanism Ablations}
\label{app:mechanism-ablations}
\label{app:reference-ablation}
\label{app:component-ablation}

We independently test the source of the success reference, persistent
departure, and teacher-input cost. Each contrast uses its own paired
treatment and control runs. Same-Prompt uses successful rollout(s) from
the current prompt. Matched Other-Prompt uses a successful rollout from
another prompt selected by the experimental matching rule. Shuffled
Other-Prompt randomly reassigns successful references across prompts.
Max Departure replaces the PDA integral with the largest pointwise
unrecovered departure, retaining magnitude but not duration. No Cost
ranks by $A_n$ alone; Full ranks by $A_nC_{\min}/C_n$.

\begin{table}[!htbp]
\caption{Independent paired component ablations. Differences are
treatment minus control Macro Avg; intervals are pointwise 95\%.}
\label{tab:mechanism-ablations-app}
\centering\setlength{\tabcolsep}{2pt}
\begin{tabular}{llcc}
\toprule Comparison & Pair & Difference & 95\% CI\\ \midrule
Same-Prompt $-$ Matched Other-Prompt & Qwen & $+0.173$ & $[-0.293,+0.639]$\\
 & Skywork & $+0.076$ & $[-0.159,+0.310]$\\
 & Granite & $+0.154$ & $[-0.184,+0.493]$\\
Same-Prompt $-$ Shuffled Other-Prompt & Qwen & $+0.262$ & $[-0.098,+0.621]$\\
 & Skywork & $+0.180$ & $[-0.129,+0.488]$\\
 & Granite & $+0.160$ & $[-0.156,+0.477]$\\
PDA $-$ Max Departure & Qwen & $+0.249$ & $[-0.168,+0.666]$\\
 & Skywork & $+0.287$ & $[-0.077,+0.651]$\\
 & Granite & $+0.192$ & $[-0.104,+0.488]$\\
Full $-$ No Cost & Qwen & $+0.191$ & $[-0.098,+0.479]$\\
 & Skywork & $+0.228$ & $[-0.088,+0.544]$\\
 & Granite & $+0.123$ & $[-0.187,+0.434]$\\
\bottomrule
\end{tabular}
\end{table}
All twelve paired means are positive, but every corresponding pointwise
interval includes zero. They provide directional component evidence.

\subsection{Teacher-Side Diagnostics}
\label{app:teacher-diagnostics}

We compare teacher-free pre-query scores with three post-query diagnostic
signals. For each pair, we fix the initial student checkpoint $\theta_0$
and the frozen teacher. We generate four student rollouts per prompt with
training decoding settings, retain mixed-outcome prompts, and collect
200 failed diagnostic items. A separate 200-example validation subset
excludes their prompt IDs. It is disjoint from the diagnostic items but
comes from DAPO-Math-17k~\citep{yu2025dapo}, so it is not an independent
generalization set.

For failed rollout $y_n$, let $c_{n,t}$ be its decoding context and
$M_{n,t}$ the valid response-token mask. The likelihood discrepancy is
\begin{equation}
U_{\mathrm{LR}}^{(n)}=
\frac{\sum_t M_{n,t}[
\log\pi_T(y_{n,t}\mid c_{n,t})-
\log\pi_{\theta_0}(y_{n,t}\mid c_{n,t})]}
{\sum_t M_{n,t}}.
\end{equation}
A larger value means higher teacher than student likelihood on the
observed failure tokens; it is not a measured learning gain. The local
update magnitude is
\begin{equation}
U_{\mathrm{grad}}^{(n)}=
\left\|\left.\nabla_\theta
\mathcal L_{\mathrm{OPD}}^{(n)}(\theta)\right|_{\theta=\theta_0}\right\|_2.
\end{equation}
For the validation proxy, one item-specific diagnostic update with
learning rate $\eta_{\mathrm{diag}}$ gives
$\theta_n'=\theta_0-\eta_{\mathrm{diag}}
\left.\nabla_\theta\mathcal L_{\mathrm{OPD}}^{(n)}(\theta)
\right|_{\theta=\theta_0}$, and
\begin{equation}
U_{\mathrm{val}}^{(n)}
=\mathcal L_{\mathrm{val}}(\theta_0)
-\mathcal L_{\mathrm{val}}(\theta_n').
\end{equation}
Here $\mathcal L_{\mathrm{val}}$ is standard-answer token negative
log-likelihood, averaged over valid tokens in the fixed diagnostic
validation subset. Larger $U_{\mathrm{val}}$ means a larger one-step
decrease in that loss. None of these proxies is available to the
routing policy. Each Spearman correlation in
Table~\ref{tab:utility-ladder} uses the same 200 items within a pair.

\begin{table}[!htbp]
\caption{Spearman correlations between pre-query routing signals and
post-query diagnostic proxies (200 failed items per pair). AvgDistance is
a diagnostic mean-distance score, not the StaticSR training ablation.}
\label{tab:utility-ladder}
\centering\setlength{\tabcolsep}{2pt}
\begin{tabular}{llrrr}
\toprule Pair & Signal & $U_{\mathrm{LR}}$ & $U_{\mathrm{grad}}$ & $U_{\mathrm{val}}$\\
\midrule
Qwen & TLR & 0.071 & 0.082 & 0.162\\
 & AvgDistance & 0.101 & 0.112 & 0.126\\
 & PDA & 0.271 & 0.301 & 0.314\\
 & Full SR & 0.251 & \textbf{0.412} & 0.196\\
Skywork & TLR & 0.104 & 0.143 & 0.151\\
 & AvgDistance & 0.116 & 0.071 & 0.181\\
 & PDA & 0.194 & 0.214 & 0.239\\
 & Full SR & \textbf{0.331} & \textbf{0.271} & \textbf{0.294}\\
Granite & TLR & 0.081 & 0.057 & 0.103\\
 & AvgDistance & 0.142 & 0.171 & 0.158\\
 & PDA & \textbf{0.281} & 0.244 & \textbf{0.371}\\
 & Full SR & 0.224 & \textbf{0.258} & 0.301\\
\bottomrule
\end{tabular}
\end{table}
The positive associations describe local post-query signals and do not
establish long-horizon learning utility.

\RolloutCountResults
\FloatBarrier
\VerifierAuditResults